\documentclass[10pt,twocolumn,letterpaper]{article}

\usepackage{cvpr}
\usepackage{times}
\usepackage{epsfig}
\usepackage{graphicx}
\usepackage{amsmath}
\usepackage{amssymb}
\usepackage{url}            % simple URL typesetting
\usepackage{booktabs}       % professional-quality tables
\usepackage{amsfonts}       % blackboard math symbols
\usepackage{nicefrac}  
\usepackage{enumitem}
\usepackage{lipsum}
\usepackage{microtype}      % microtypography
\usepackage{graphicx}
\usepackage{multicol}
\usepackage{wrapfig,subfig}
\usepackage{subfig}
\usepackage{color}
\usepackage{float}
\usepackage{bm}
\usepackage{dashrule}
\usepackage[table]{xcolor}

\usepackage[]{algorithm2e}

\DeclareMathOperator*{\argmin}{arg\,min}

\newcommand{\salt}[0]{SALT }

\def\etal{\emph{et. al.}}

\def\ie{\emph{i.e}\onedot}

\newcommand{\norm}[1]{\left\lVert#1\right\rVert}

\usepackage[pagebackref=true,breaklinks=true,letterpaper=true,colorlinks,bookmarks=false]{hyperref}

\cvprfinalcopy % *** Uncomment this line for the final submission

\def\cvprPaperID{8480} % *** Enter the CVPR Paper ID here

\ifcvprfinal\fi
\begin{document}

%%%%%%%%% TITLE
\title{\salt: Subspace Alignment as an Auxiliary Learning Task for Domain Adaptation}

\author{Kowshik Thopalli$^{\dagger}$, Jayaraman J. Thiagarajan$^{\ddagger}$, Rushil Anirudh$^{\ddagger}$, Pavan Turaga$^{\dagger}$\\
$^{\dagger}$Arizona State University, $^{\ddagger}$Lawrence Livermore National Labs\\
}

\maketitle
%\thispagestyle{empty}

%%%%%%%%% ABSTRACT
\begin{abstract}
 Unsupervised domain adaptation aims to transfer and adapt knowledge learned from a labeled source domain to an unlabeled target domain. Key components of unsupervised domain adaptation include: (a) maximizing performance on the target, and (b) aligning the source and target domains. Traditionally, these tasks have either been considered as separate, or assumed to be implicitly addressed together with high-capacity feature extractors. When considered separately, alignment is usually viewed as a problem of aligning data distributions, either through geometric approaches such as subspace alignment or through distributional alignment such as optimal transport. This paper represents a hybrid approach, where we assume simplified data geometry in the form of subspaces, and consider alignment as an auxiliary task to the primary task of maximizing performance on the source. The alignment is made rather simple by leveraging tractable data geometry in the form of subspaces. We synergistically allow certain parameters derived from the closed-form auxiliary solution, to be affected by gradients from the primary task. The proposed approach represents a unique fusion of geometric and model-based alignment with gradients from a data-driven primary task. Our approach termed SALT, is a simple framework that achieves comparable or sometimes outperforms state-of-the-art on multiple standard benchmarks.
\end{abstract}

%%%%%%%%% BODY TEXT
 \section{Introduction}
Despite significant advances in neural network architectures and optimization strategies for supervised learning, one of the long-standing challenges has been to effectively generalize classifier models to novel testing scenarios, typically characterized by unknown covariate shifts~\cite{judyiclr13} or changes in label distributions. In this paper, we consider the problem of \textit{unsupervised domain adaptation}, wherein the goal is to utilize labeled data from a \textit{source} domain to design a classifier that can generalize to an unlabeled \textit{target} domain. We are especially interested in the case when no knowledge about the covariate shift is available. The covariate shift commonly considered in unsupervised domain adaptation formulations assume that the distributions on source and target domains differ only in their marginal feature distributions $P(X)$ while having an identical conditional distribution $P(y|X)$, where $X$ and $y$ are correspond to features and labels from either the source $(\mathrm{X}_s, \mathrm{y}_s)$ or target $(\mathrm{X}_t, \mathrm{y}_t)$ domains.

More successful solutions for domain adaptation attempt to infer domain-invariant data representations by directly minimizing the discrepancy between the marginal feature distributions from the two domains. For example, domain adversarial learning, which seeks to find a common representation where the two domains are indistinguishable, is at the core of several state-of-the-art  methods~\cite{tzeng2017adversarial,CYCADA,CDAN,cite:ICML15RevGrad,cite:JMLR16RevGrad}. However, it has recently been shown that domain adversarial training can be ineffective when working with a high-capacity feature extractor \cite{shu2018a}. High-capacity networks allow for learning arbitrary transformations that can reduce domain mismatch in terms of marginal feature distributions, yet might have no bearing on the final classifier performance \cite{shu2018a}. This has motivated the inclusion of a variety of consistency-enforcing losses into the domain adversarial learning formulation to regularize the learning process. For example, both feature and semantic losses for feature-level adaptation may be employed ~\cite{cite:JMLR16RevGrad,tzeng2017adversarial}, while pixel-level adaptation via pixel and semantic consistency losses may also be employed \cite{liu2016coupled,bousmalis2017unsupervised}. More recently, Hoffman \etal ~\cite{CYCADA} proposed to enforce cyclical consistency based on all the aforementioned losses, while Shu \etal ~\cite{shu2018a} introduced a virtual adversarial loss to better regularize domain adversarial learning.\\

\noindent \textbf{Key insights:} The above discussion leads us to our core idea that the process of minimizing domain discrepancies, while also learning a highly generalizable classifier, could be potentially regularized by adopting alignment methods with simplified data geometries. A natural candidate is subspace alignment~\cite{Subspace_alignment,Gong2012GeodesicFK,ShrivastavaWACV2014,thopalli2019multiple}, which utilizes simplified data representations, i.e., low-dimensional linear subspaces, and poses the problem of achieving domain invariance as learning a mapping between those representations. Despite their mathematical tractability, these methods are typically agnostic to the end-task and rely on modeling assumptions that are insufficient to describe complex datasets. Thus, they perform poorly in comparison to more recent approaches in domain adaptation. Consequently, one must attempt to blend the representational convenience of simplified data geometries, while not being constrained by the limited capabilities of these methods. 

% Analytic solutions for alignment while convenient, can cause errors due to a mismatched geometry to propagate downstream. 
To this end, we develop SALT, an unsupervised domain adaptation algorithm based on simple subspace-based alignment, which is capable of producing highly effective classifiers through synergistic optimization between improving classifier performance and minimizing domain mismatch. Intuitively, by handling the interactions between domain alignment and end-task objectives, we simultaneously regularize the domain adaptation process and eliminate the commonly observed performance limitations of subspace-based methods. \\
%In particular, SALT strikes a balance between the following factors:  a) assumes tractable data geometries in source and target domains, which can be analytically leveraged for data alignment, b) synergistically adapts certain parameters derived from the analytic solution to alignment, in a manner that maximizes performance on the primary task of classification. 

\noindent \textbf{Contributions and findings}:
In this paper, we cast explicit domain alignment as an {\em auxiliary} task, whose fidelity can be carefully adjusted to maximize the quality of the primary task, i.e., performance of the classifier on both source and target domains. More specifically, we define {\em adaptable subspace alignment} as the auxiliary task, which uses gradients from the primary task, to adjust the domain alignment. We show that even with a simplified global subspace alignment model, SALT yields a comparable or sometimes higher adaptation performance than even state-of-the-art methods with sophisticated adaptation strategies. %This approach falls under non-conservative adaptation, and hence we include explicit information-invariance losses for the unlabeled target domain, similar to~\cite{shu2018a}.
Our major findings are: 
\begin{itemize}[noitemsep]
    
      \item With a disjoint primary-auxiliary formulation, we find that even a na\"ive global subspace based alignment~ \cite{Subspace_alignment} with a \textit{fixed} feature extractor, achieves higher or similar performance compared to state-of-the-art  approaches on several benchmarks.
      
      \item By viewing domain alignment as an auxiliary task, we are able to entirely dispense the need for adversarial learning, consistency-enforcing regularizers, and other extensive hyper-parameter choices.
      
      \item We find SALT to be robust to varying data availability in the target domain, which can be attributed to the simplified data representations used for alignment.
      
      \item Though SALT uses only linear subspaces for alignment, we are able to increase its complexity through the use of an ensemble of subspace models and achieve improved performance.
\end{itemize}

 \section{Related work}
\begin{figure*}[t]
	\centering
	\centerline{\includegraphics[width=1\linewidth]{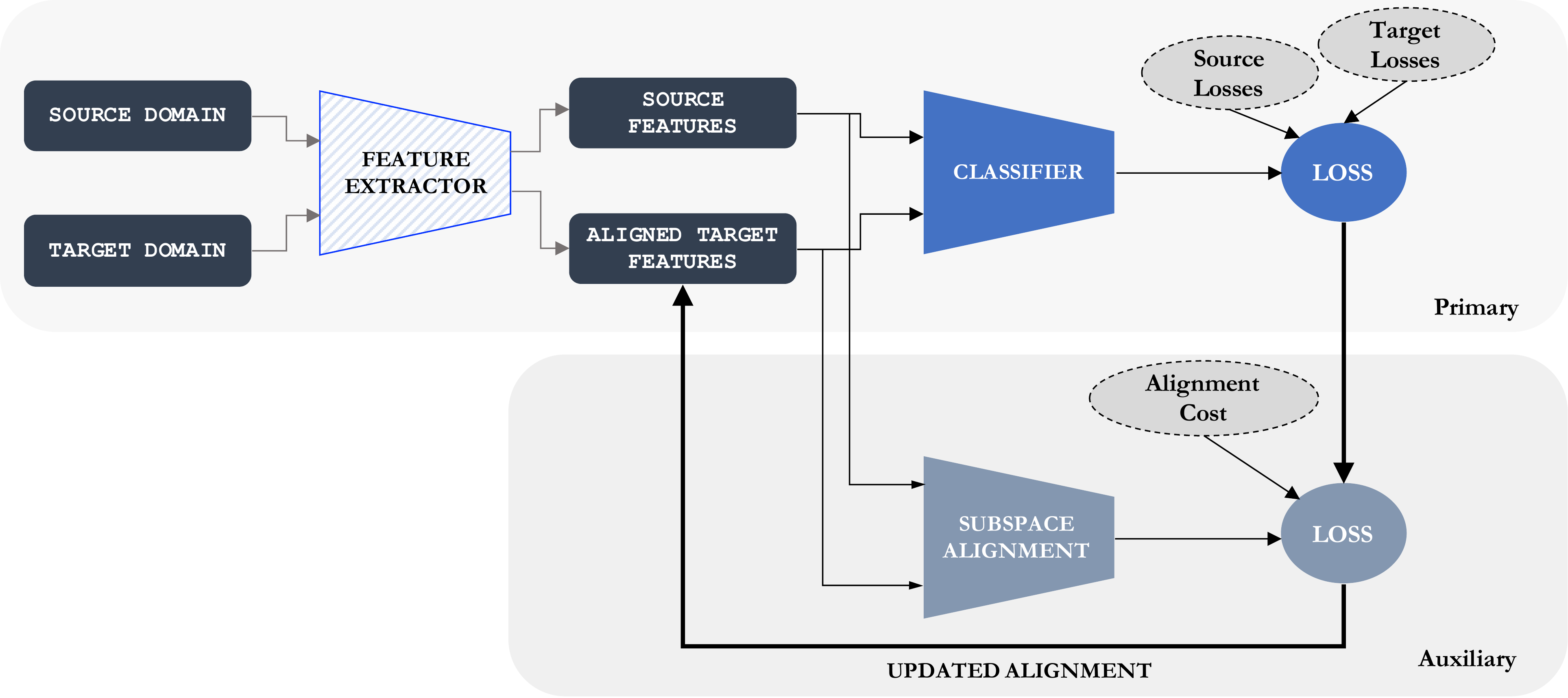}}
	% \vspace{-10pt}
	%  \vspace{2.0cm}
	\caption{\small{An overview of the proposed approach for unsupervised domain adaptation. We leverage gradients from the \textit{primary task} of designing a generalizable classifier to guide the domain alignment, which is posed as an \textit{auxiliary task}. While the primary task utilizes deep neural networks, the auxiliary task is carried out using a simplified data geometry -- subspaces --  in lieu of adversarial training or sophisticated distribution matching. Note, even the \textit{feature extractor} is frozen after an initial training phase.}}
	\label{fig:Proposed approach}
\vspace{-0.2in}
\end{figure*}
\label{relatedwork}
%In this section, we briefly review the prior art in unsupervised domain adaptation. We will also discuss about meta auxiliary learning, which is closely related to the proposed approach.

\noindent \textbf{Unsupervised domain adaptation}:
Unsupervised domain adaptation has been an important problem of research in multiple application areas and a wide variety of solutions have been developed. Earlier works focused on adapting the features of source and target domains by minimizing statistical divergence between them \cite{Saenko2010AdaptingVC,Gong2012GeodesicFK,pan2010domain,sun2017correlation,Fernando2014SubspaceAF,sun2015subspace}. These works can be analyzed through the foundational work of Ben David \etal ~\cite{Ben-David:2010:TLD:1745449.1745461}, which provides an upper bound on target error, $\epsilon(\mathcal{D}_T; h)$ on target data $\mathcal{D}_T$, that can be achieved using a hypothesis $h$ as the sum of three terms: 
\begin{equation}
\epsilon(\mathcal{D}_T; h) \leq \mathcal{L}(\mathcal{D}_S; h)+  \mathcal{L}_{\mathcal{H}} (\mathcal{D}_S, \mathcal{D}_T)+ \mathcal{L}_{\delta}(h),
\label{eq:error}
\end{equation}
where, the first term denotes the error in the source domain $\mathcal{S}$, the second term is the discrepancy between the source-target pair ($\mathcal{H}$-divergence), and the third term measures the optimal error achievable in both the domains (often assumed to be negligible). Under this context, there are two broad categories of methods -- ones that assume there exists a single hypothesis $h$ that can perform well in both domains (\textit{conservative}), and those that do not make that assumption (\textit{non-conservative})~\cite{shu2018a}. Successful state-of-the art methods use powerful feature extractors such as convolutional neural networks (CNNs), and aim to jointly minimize source error along with domain divergence error. Adversarial learning~\cite{cite:NIPS14GAN} has been the workhorse of these solutions, implemented with different additional regularizers~\cite{cite:JMLR16RevGrad,CDAN,CYCADA, UNIT,cite:CVPR17pix2pix}.

\noindent \textbf{Subspace-based alignment}:
The key idea behind this class of methods is to represent source and target data distributions on lower-dimensional subspaces, align the subspaces, and subsequently project the target data onto the aligned subspace. A classifier is finally trained on the newly computed lower dimensional source data and evaluated on target data.  Popular approaches include~\cite{Gong2012GeodesicFK,Gopalan2011DomainAF,Subspace_alignment,sun2015subspace}. Geodesic-based methods \cite{Gopalan2011DomainAF,Gong2012GeodesicFK} compute a path along the manifold of subspaces (Grassmannian), and either project the
source and target onto points along that path~\cite{Gopalan2011DomainAF} or compute a linear map that projects source samples directly onto the target subspace~\cite{Gong2012GeodesicFK}. 
Furthermore, works such as \cite{Subspace_alignment,sun2015subspace} align the source and target subspaces using Procrustes methods \cite{Subspace_alignment}, or by considering distributional statistics along with subspace basis \cite{sun2015subspace}.

\noindent \textbf{Meta auxiliary learning}:
Meta-learning has been a recently successful approach in generalizing knowledge across related tasks~\cite{maml}. Broadly, meta-learning techniques can be grouped into three categories~\cite{maml} -- metric-based~\cite{koch2015siamese,vinyals2016matching}, model-based~\cite{Santoro:2016:MMN:3045390.3045585,MunkhdalaiY17} and optimization-based~\cite{maml,RaviL17}. Auxiliary learning on the other hand essentially focuses on increasing the performance of a primary task through the help of another related auxiliary task(s). This methodology has been applied to areas such as speech recognition~\cite{ToshniwalTLL17}, depth estimation, semantic segmentation~\cite{Liebel}, and reinforcement learning~\cite{JaderbergMCSLSK17}. The work closely related to ours is \textit{meta-auxiliary learning} \cite{metaauxillary}, which aims to improve $m$-class image classification performance (primary task) by solving a $k$-class classification problem (auxiliary task). This is done by establishing a functional relationship between the classes. In contrast, we formulate subspace-based domain alignment as the auxiliary to the primary task of building a classifier that works well in both source and target domains.

 \section{Proposed Approach}
\label{methods}
In this section, we describe the proposed method for unsupervised domain adaptation. 
An overview of the approach can be found in Figure \ref{fig:Proposed approach}. 
We assume access to data from the labeled source and unlabeled target domains, $\mathcal{D}_S $ and  $\mathcal{D}_T$ denoted as $\{\mathrm{X}_s,\mathrm{y}_s\} $ and $\{\mathrm{X}_t\} $ respectively. In the rest of this paper, we use $\mathrm{X}_s, \mathrm{X}_t$ to indicate the latent features for source and target domains from a pre-trained feature extractor, $\mathcal{F}$, such as ResNet50 \cite{ResNet}. 
The primary network updates the classifier given the source and current best estimate of source-aligned target features, such that the inferred model is effective for both source and target domains. The auxiliary network solves for subspace-based domain alignment, while minimizing both the alignment cost, and the loss from the primary network. Though the resulting alignment is sub-optimal in terms of the pure alignment cost, it is optimal when conditioned on the primary classification task. 

\subsection{\textit{Primary Task}: Classifier design}
\label{sec:primary}
We construct the primary task with the goal of achieving effective class discrimination in both source and target domains. 
With inputs as source/target images directly, or latent features extracted from a pre-trained feature extractor $\mathcal{F}$, we learn the parameters for a classifier network $\mathcal{P}_{\Theta}$ parameterized by $\Theta$. 
The losses used for the optimization include: (i) standard categorical cross-entropy loss $\mathcal{L}_{y}\left(\Theta ; \mathcal{D}_{S}\right)=\mathbb{E}_{x, y \sim \mathcal{D}_{S}}\left[y^{\top} \ln \mathcal{P}_{\Theta}(x)\right]$ for the labeled source data, (ii) conditional entropy~\cite{shu2018a} loss on the softmax predictions for target data $\mathcal{L}_{c}\left(\Theta ; \mathcal{D}_{T}\right)=-\mathbb{E}_{x \sim \mathcal{D}_{T}}\left[\mathcal{P}_{\Theta}(\mathcal{A}_{\Phi}(x))^{\top} \ln \mathcal{P}_{\Theta}(\mathcal{A}_{\Phi}(x))\right]$, and (iii) class-balance loss~\cite{FrenchMF18} for the unlabeled target domain $\mathcal{L}_{cb}$, which is implemented as binary cross-entropy loss between the mean prediction from the network over a mini-batch to that of a uniform probability vector -- this loss regularizes network behavior when the data exhibits large class imbalance. Note that, in the definition of $\mathcal{L}_{c}$, the target domain features are first transformed using the auxiliary network $\mathcal{A}_{\Phi}$ (defined in Section \ref{sec:SA-network}) prior to applying the classifier. The overall loss function is thus defined as 
\begin{equation}
\label{eqn:sum losses}
    \mathcal{L}_{\mathcal{P}}=\mathcal{L}_{y}+\lambda_c \mathcal{L}_{c}+\lambda_{cb}\mathcal{L}_{cb}. 
\end{equation}Here, the second and third terms are used as regularizers to counter the assumption that a single hypothesis $h$ might not be effective for both domains, \ie non-conservative. 

% In its simplest form, this formulation should work if there is no covariate shift between the domains. 
% However, in our setup, in order to account for unknown shifts (if they exist), we formulate an auxiliary task for domain alignment.

% {\color{red}{Have to mention here that the target data is transformed 
% by our auxiliary design}}

\subsection{\textit{Auxiliary Task}: Domain alignment}
\label{sec:SA-network}
We posit that an alternating optimization between a generalizable classification task, and an auxiliary domain alignment task, relaxes the requirements of the alignment step such that even simple alignment strategies can provide sufficient information to improve the classifier.
In order to test this idea, we assume a simplified data geometry, in the form of low-dimensional linear subspaces~\cite{Subspace_alignment}. 
Note that, as a generative model for a dataset, a single linear subspace or even a union of linear subspaces is a poor choice on its own. However, when coupled with an appropriate primary task using a sufficiently high capacity classifier, we will show it can be highly effective in domain adaptation. Though we report results only with subspace-based domain alignment, without loss of generality, the same algorithm can be extended to other domain alignment approaches including domain adversarial training~\cite{DAN}, optimal transport~\cite{Courty2017OptimalTF} etc.

% {\color{red}{write about other sophisticated methods}}

\noindent \textbf{Closed-form subspace alignment:} Let us denote the basis vectors for the $d$-dimensional subspaces inferred from source and target domains as $\{\mathrm{Z}_s\}$ and $\{\mathrm{Z}_t\}$  respectively and they satisfy 
$\mathrm{Z}_s^T\mathrm{Z}_s = \mathbb{I}$, $\mathrm{Z}_t^T\mathrm{Z}_t = \mathbb{I}$, where $\mathbb{I}$ denotes the identity matrix. 
The subspaces are inferred using singular value decomposition of source/target domain latent features $\{\mathrm{X}_s, \mathrm{X}_t\}$.
The alignment between two subspaces can be parameterized as an affine transformation $\mathrm{\Phi}$, \ie
\begin{align}
    \label{eq:SAobjective}
    \begin{split}
    \mathrm{\Phi}^* = \argmin_\mathrm{\Phi} \norm{\mathrm{Z}_t \mathrm{\Phi} - \mathrm{Z}_s}_F^2,
    \end{split}
\end{align}where, $\norm{.}_F$ denotes the Frobenius norm. The solution to this alignment cost \eqref{eq:SAobjective} can be obtained in closed-form \cite{Subspace_alignment} as
\begin{equation}
    \label{eq:globalsoln}
        \mathrm{\Phi}^* = (\mathrm{Z}_t)^T \mathrm{Z}_s.
\end{equation}This implies that the adjusted coordinate system, also referred as the \textit{source-aligned target subspace} can be constructed as 
 \begin{equation} \mathrm{Z}_{t}^a =  \mathrm{Z}_t (\mathrm{Z}_t)^T \mathrm{Z}_s.
    \label{eqn:align}
\end{equation}
Though we develop our formulation by aligning the target subspace onto the source, without loss of generality, one can equivalently project the source subspace onto the target. Since the primary task invokes the classifier optimization using features in the ambient space, we need to re-project the target features using $\mathrm{Z}_{t}^a$, \ie
\begin{align}
{\hat{\mathrm{X}}_t}^* &=\argmin_{\hat{\mathrm{X}}_t} \norm{\hat{\mathrm{X}}_t \mathrm{Z}_s -\hat{\mathrm{X}}_{t} \mathrm{Z}_t^a}_{F}^2 \nonumber \\
&=\argmin_{\hat{\mathrm{X}}_t} \norm{\hat{\mathrm{X}}_t \mathrm{Z}_s -\hat{\mathrm{X}}_{t}  \mathrm{Z}_t (\mathrm{Z}_t)^T \mathrm{Z}_s}_{F}^2,
\end{align}where $\hat{\mathrm{X}}_t^*$ denotes the modified target features. The solution to this optimization can be obtained in closed-form as
\begin{align}
    \label{eq:reproj}
    \mathcal{A}_{\Phi}(\mathrm{X}_t) = {\hat{\mathrm{X}}_t}^* &= \mathrm{X}_t\mathrm{Z}_t\mathrm{\Phi}^*\mathrm{Z}_s^T,
\end{align}where $\mathrm{\Phi}^*$ is computed using \eqref{eq:globalsoln}.

 \RestyleAlgo{boxruled}
\begin{algorithm}[t]
	\textbf{Input}: Labeled source features $\{\mathrm{X}_s,\mathrm{y}_s\} $ and 
	unlabeled target features $\{\mathrm{X}_t\}$ from $\mathcal{F}$. Source and target subspaces $\mathrm{Z}_s$;$\mathrm{Z}_t$ 
	
	\textbf{Initialize}: Random state for $\Theta$, Alignment $\mathrm{\Phi}$ using \eqref{eq:globalsoln}.
	Hyper-parameters $\lambda_c,\lambda_{cb},\gamma_c,\gamma_{cb}, n_{iter},T_1,T_2 $.\\
	\noindent \textbf{Training Phase}:\\
     Split: $\mathrm{X}_S^{\dagger} , \mathrm{X}_S^{\ddagger} \leftarrow  \mathrm{X}_S$ and $\mathrm{X}_T^{\dagger} , \mathrm{X}_T^{\ddagger} \leftarrow \mathrm{X}_T$  \\
 	\For{iter \textbf{in} $n_{iter}$}{
 	     \tcp{ update $\mathcal{P}_{\Theta}$}
 	    \For{$t_1$ \textbf{in} $T_1$}{
	    		Compute $\bm\hat{\mathrm{X}}_t^{\dagger} = \mathrm{X}_t^{\dagger}\mathrm{Z}_t\mathrm{\Phi}^*\mathrm{Z}_s^T $ following \eqref{eq:reproj}\;
			$\bm\hat{\mathrm{y}}_s^{\dagger} = \mathcal{P}_{\Theta}(\bm\hat{\mathrm{X}}_s^{\dagger})$\;
			$\bm\hat{\mathrm{y}}_t^{\dagger} = \mathcal{P}_{\Theta}(\bm\hat{\mathrm{X}}_t^{\dagger})$\;
			
			Compute $\mathrm{L}_{\mathcal{P}}$ using \eqref{eqn:sum losses}\;
		
			Update $\Theta^* = \arg \min_{\Theta} \mathrm{L}_{\mathcal{P}} $ \;}
            
	        \tcp{ update $\mathcal{A}_{\Phi}$}
			\For{$t_2$ \textbf{in} $T_2$}{

			Compute $\bm\hat{\mathrm{X}}_t^{\ddagger} $ using \eqref{eq:reproj}\;
			Compute $\bm\hat{\mathrm{y}}_t^{\ddagger} =\mathcal{P}_{\Theta^*}(\bm\hat{\mathrm{X}}_t^{\ddagger})  $;\\

			Compute $\mathcal{L}_{\mathcal{A}}$ using \eqref{eqn:loss_A}\;
			Update $\mathrm{\Phi}^* = \arg \min_{\Phi} \mathrm{L}_{\mathcal{A}_{\Phi}} $ \;
			
			}
		}

	\caption{\small{\salt for unsupervised domain adaptation.}}
	\label{algo:algo}
\end{algorithm}

\noindent \textbf{Task-dependent tuning of subspace alignment: }
Since the overall objective is to refine the auxiliary network parameters to maximally support the primary task, we propose to include the terms $\mathcal{L}_c$ and $\mathcal{L}_{cb}$ from \eqref{eqn:sum losses} to the alignment objective in \eqref{eq:SAobjective}, 
\begin{equation}
    \mathcal{L}_{\mathcal{A}} = \norm{\mathrm{Z}_t \mathrm{\Phi} - \mathrm{Z}_s}_F^2 + \gamma_c \mathcal{L}_{c}+\gamma_{cb}\mathcal{L}_{cb}.
    \label{eqn:loss_A}
\end{equation}
Note that, when we make this modification, there no longer exists a closed-form solution. Hence, we adopt an approach that takes in gradients from the primary task to adjust $\mathrm{\Phi}$. To enable this end-to-end training of both the primary and auxiliary tasks, we implement \textit{subspace alignment} as a network $\mathcal{A}$ that parameterizes $\mathrm{\Phi}$ as a fully connected layer of $d$ neurons without any non-linear activation function or bias.

\begin{table*}[h]
    %\large
    \begin{center}
    \begingroup
    %\setlength{\tabcolsep}{10pt} % Default value: 6pt
    %\renewcommand{\arraystretch}{1.2}
 %   \resizebox{\columnwidth}{!}{%
 \renewcommand*{\arraystretch}{1.2}
    \begin{tabular}{cccccccc}
    
    \hline
    %\cellcolor{gray!25}{\textbf{Domains}} & \cellcolor{gray!25}\textbf{NA} & \cellcolor{gray!25}\textbf{GFK}& \cellcolor{gray!25}\textbf{SA}& \cellcolor{gray!25}\textbf{SDA} & \cellcolor{gray!25}\textbf{PT} & \cellcolor{gray!25}\textbf{Proposed} \\

 Method & I $\rightarrow$ P & P $\rightarrow$ I & I $\rightarrow$ C & C $\rightarrow$ I & C $\rightarrow$ P & P $\rightarrow$ C & Average \\ 
\hline
No Adaptation & 76.5 & 88.2 & 93 & 84.3 & 69.1 & 91.2 & 83.7 \\
DAN~\cite{DAN} & 74.5 &82.2 & 92.8 & 86.3 & 69.2 & 89.8 & 82.5 \\
DANN~\cite{cite:JMLR16RevGrad} & 75.0 & 86.0 & 96.2   & 87.0 &74.3 &91.5 &85.0 \\
JAN~\cite{JAN} & 76.8 &88.0& 94.7& 89.5& 74.2 & 91.7& 85.8 \\

CDAN+E~\cite{CDAN} & {\bf \em 78} & {\bf \em 90.9} & \textbf{98.1} & \textbf{91.6} & {\bf \em 74.4} & {\bf \em 94.6} & {\bf \em 87.9} \\
\hline
SALT & \textbf{80.16} & \textbf{95.5} & {\bf \em 97.3} & {\bf \em 90.9} & \textbf{79.3} & \textbf{97} & \textbf{90.02} \\

    \hline
\end{tabular}
%}
  
    \endgroup
    \caption{\small{Classification accuracy on the ImageCLEF dataset. Best performance is shown in {\bf bold}, and the second best in {\em \bf \em bold italic}.}}
    \label{table:image_clef}
    \end{center}
    \end{table*}

\noindent \textbf{Objective:} The overall objective of this primary-auxiliary network learning can be formally written as the following bi-level optimization problem:
\begin{align}
&\min_{\Theta} \mathcal{L}_{\mathcal{P}}\left(\Theta; \mathrm{X}_s,\mathrm{y}_s, \mathcal{A}_{\Phi^*}(\mathrm{X}_t)\right), \\
\nonumber \text{where,} \quad \mathrm{\Phi}^* = &\arg \min_{\mathrm{\Phi}} \mathcal{L}_{\mathcal{A}}\bigg(\mathrm{\Phi}; \mathrm{Z}_s,\mathrm{Z}_t,  \mathcal{P}_{\Theta}(\mathcal{A}_{\Phi}(\mathrm{X}_t))\bigg).
\label{eqn:obj}
\end{align}We now describe the algorithm for solving this objective. 
% The auxiliary network $\mathcal{A}_{\Phi}$ implements the function derived in \eqref{eq:reproj} \ie $\mathcal{A}$ transforms target data through the source aligned target subspace and re-projects it into the ambient space.  

% Where $\mathcal{L}_{\mathcal{A}}$ represent the linear combination of the primary classification loss on source aligned target data $\hat{\mathrm{X}}_t$ (equation \ref{eq:reproj}) along with subspace alignment cost given by equation \ref{eq:SAobjective}. 

\subsection{Algorithm}
Given the primary and auxiliary task formulations, one can adopt different training strategies to combine their estimates: (i) \textit{Independent}: This is the classical approach, where the alignment obtained by solving \eqref{eq:SAobjective} is used to infer the classifier parameters, (ii) \textit{Joint}: This jointly optimizes for both networks together, similar to existing domain adaptation methods, (iii) \textit{Alternating}: This alternating style of optimization solves for the primary task with the current estimate of the alignment, and subsequently updates the auxiliary network with both primary and auxiliary losses. As we will show later, that this  alternating optimization strategy works the best in comparison to the other two. We now describe the alternating optimization strategy.

\noindent \textbf{Initialization phase}: The choice of initial states for the parameters of both the primary and auxiliary networks is crucial to the performance of our algorithm. First, we pre-train the feature extractor $\mathcal{F}$ and the classifier $\mathcal{P}_{\Theta}$ using the loss function \eqref{eqn:sum losses} without any explicit domain alignment. We then fit $d$-dimensional subspaces, $\mathrm{Z}_s$ and $\mathrm{Z}_t$, to the features obtained using $\mathcal{F}$ for both the source and target domains. 
Note that the feature extractor is not updated for the rest of the training process, and hence the subspace estimates are fixed. 
The initial state of $\mathrm{\Phi}$, \ie alignment matrix between the two subspaces, is obtained using \eqref{eq:globalsoln}.

\noindent \textbf{Training phase}: In order to enable information flow between the two tasks, we propose to allow the auxiliary task to utilize gradients from the primary task. Similarly, the estimated alignment is applied to the target data while updating the classifier parameters in the primary task. The auxiliary loss construction described in Section \ref{sec:SA-network} provides a link between the primary and auxiliary tasks. 

The primary and auxiliary tasks are solved alternatively until convergence -- during the auxiliary task optimization, we freeze the classifier parameters and update $\mathcal{A}_{\Phi}$ using equation (\ref{eqn:loss_A}). Since the feature extractor $\mathcal{F}$ is fixed, there is no need to recompute the subspaces. In our implementation we find that optimizing the auxiliary task using a held-out validation set, distinct from that used for the primary task, leads to significant performance improvements. Given the estimate for $\mathrm{\Phi}$, we freeze the auxiliary network $\mathcal{A}_{\Phi}$ and update the classifier network using source features and source-aligned target features to minimize the primary loss in \eqref{eqn:sum losses}. Upon convergence (typically within $5-10$ iterations on all datasets considered), optimal values for both $\mathrm{\Phi}$ and $\mathcal{P}_{\Theta}$ are returned. A detailed listing of this process is provided in algorithm \ref{algo:algo}.

%  Following the model-agnostic meta learning (MAML), we could perform the meta optimization using \textit{gradients-through-gradients}. 
%  However, even without that, our approach produces highly effective generalization on all benchmark datasets.

\noindent \textbf{Using an ensemble of subspaces}: The fidelity of the auxiliary task relies directly on the quality of the subspace approximation. 
For complex datasets, a single low-dimensional subspace is often a poor approximation. Hence, we propose to allow the complexity of the auxiliary model to be adjusted through the use of multiple target subspaces. 
To this end, we obtain independent bootstraps of the target data and fit a single low-dimensional subspace of dimension $d$ to each of them. While solving for the auxiliary task, we compute individual alignment matrices to the source with respect to the same classifier $\mathcal{P}_{\Theta}$. During the update of the classifier $\mathcal{P}_{\Theta}$, we pose this as a multi-task learning problem, wherein a single classifier is used with different source-aligned targets. This is valid since all (bootstrapped) subspaces are in the same ambient feature space. 
During test time, we treat the predictions obtained using features from different alignment matrices as an ensemble and perform majority voting for making the final prediction.

 \section{Experiments}
\label{experiments}
We evaluated the proposed method on four widely used visual domain adaptation tasks -- digits, ImageCLEF, VisDA-2017 challenge, and Office-Home datasets, and present comparisons to several state-of-the-art domain adaptation techniques. Across all the experiments, an 80-20 random split of source and target training data was performed to update the primary and auxiliary tasks.  
All experiments were run using the PyTorch framework ~\cite{paszke2017automatic} with a Nvidia-TitanX GPU.
%Hyper-parameters are chosen according to the highest target validation accuracy

\subsection{ImageCLEF-DA}
\label{sec-imageclef}
\noindent \textbf{Dataset}: ImageCLEF\footnote{\url{http://imageclef.org/2014/adaptation}} is organized by selecting common categories of images shared by three public image datasets (domains):  \textit{ImageNet ILSVRC 2012} (\textbf{I}), \textit{Caltech-256} (\textbf{C}), and \textit{Pascal VOC 2012} (\textbf{P}). There are $12$ categories, with $50$ images each,  resulting in a total of $600$ images in each domain. We conduct $6$ experiments by permuting the $3$ domains : \textbf{I} $\rightarrow$ \textbf{P}, \textbf{P} $\rightarrow$ \textbf{I}, \textbf{I} $\rightarrow$ \textbf{C}, \textbf{C} $\rightarrow$ \textbf{I}, \textbf{C} $\rightarrow$ \textbf{P}, \textbf{P} $\rightarrow$ \textbf{C}.

\begin{figure*}[t!]
    \centering
    \subfloat[Ablation study]{{\includegraphics[width=0.4\linewidth]{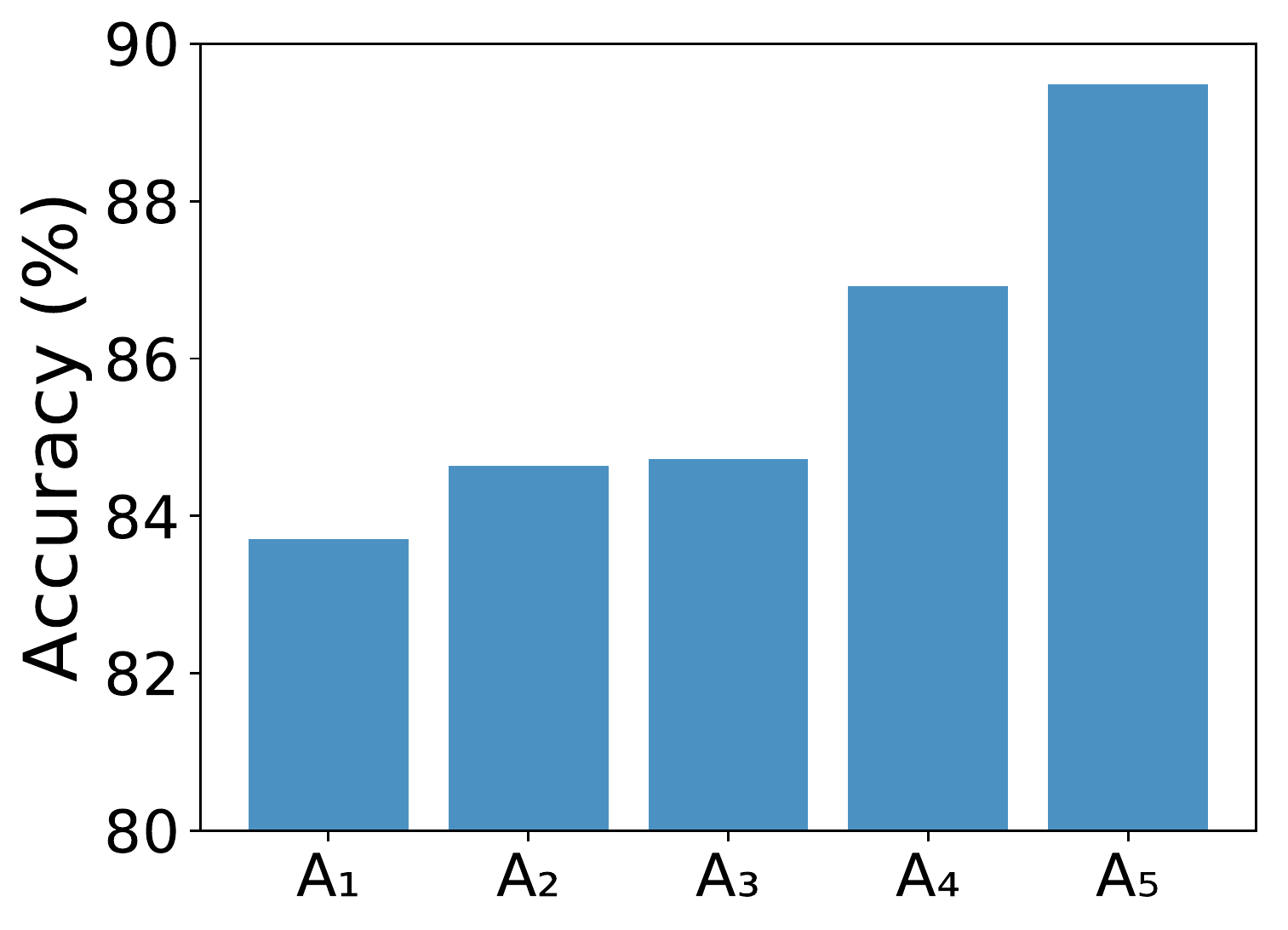} }}%
    \qquad
   % \newline
    \centering
    \subfloat[Dynamics of $\mathrm{\Phi}$ across iterations]{{\includegraphics[height =5cm,
    width=0.47\linewidth]{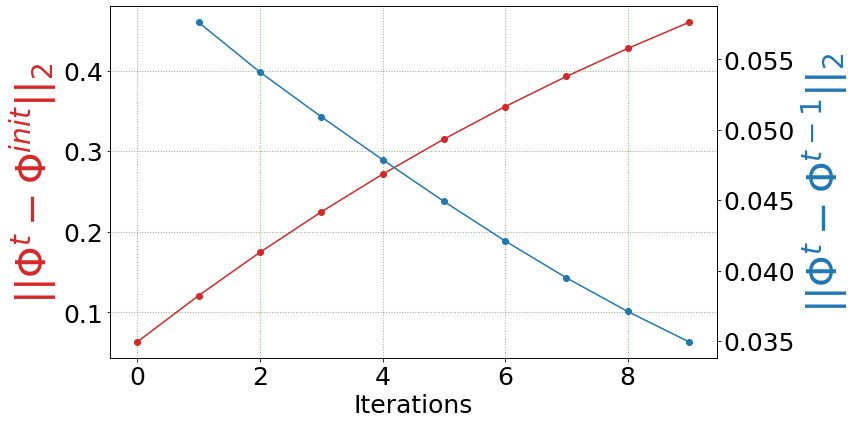} }}%
    \caption{(a) Ablating different components in the proposed method against adaptation performance on the ImageCLEF dataset. See text in Section \ref{sec-imageclef} for notation.
    (b) Changes in $\mathrm{\Phi}$ from  $\mathrm{\Phi^{init}}$ \eqref{eq:globalsoln}~ across iterations are represented by the red Curve while the blue curve denotes successive differences in $\mathrm{\Phi}$}%
    \label{fig:ablation}%
    %\vspace{-0.3in}
\end{figure*}

\noindent \textbf{Model}: Our feature extractor is based on the pre-trained ResNet-50 architecture~\cite{ResNet,russakovsky2015imagenet}. 
This model is then fine-tuned w.r.t loss computed from \eqref{eqn:sum losses}, with $\mathrm{\Phi} = \mathbb{I}$, where $\mathbb{I}$ is identity, and  $\lambda_c$ and $\lambda_{cb}$ set at $0.1$. We then use SALT on the latent features from the penultimate layer of the fine-tuned ResNet. Source and target subspaces of dimension $800$ are constructed from these $2048$-dimensional features using SVD. The classifier network is chosen to be the last fully connected layer, subsequently refined using the SGD optimizer with the learning rate 1e-4 and momentum 0.9. The subspace alignment network is trained with a learning rate of 1e-3 using the Adam optimizer 
\cite{ADAM}. 
The proposed approach is compared against a number of baseline methods including \cite{CDAN,DAN,cite:JMLR16RevGrad,JAN} and the results are reported in Table \ref{table:image_clef}. The results clearly show that even a na\"ive global alignment strategy improves performance by nearly 3 percentage points over sophisticated adversarial learning methods, with SALT's alternating optimization strategy.

\noindent \textbf{Ablation Study}: In order to understand the impact of the different components, we perform an ablation study on this dataset. We describe each setting in this experiment next: 
\begin{itemize}[noitemsep]
    \item[$\mathbf{A_1}$]  \emph{No Adaptation}: A baseline method where we use the classifier trained on the source directly on the target features without any adaptation. 
    \item[$\mathbf{A_2}$]  \emph{Primary Only}: We leave out the auxiliary task, but include all the losses used in the primary task described in equation \eqref{eqn:sum losses}, with $\mathcal{A}_{\Phi} = \mathbb{I}$.
    \item[ $\mathbf{A_3}$] \emph{Independent}: Here, we use the closed form solution in subspace alignment from equation \eqref{eqn:align}, and then solve for the primary task in \eqref{eqn:sum losses} independently. 
    \item[ $\mathbf{A_4}$] \emph{Joint Optimization}: We employ a joint optimization strategy, wherein we jointly update the alignment $\mathrm{\Phi}$, and the classifier together. 
    \item[$\mathbf{A_5}$]  \emph{Alternating Optimization}: This is our proposed strategy that updates $\mathrm{\Phi}$ and the classifier in an alternating fashion. 
    %The updates to the alignment are guided by the gradients from the primary task.
\end{itemize}The results from the study are illustrated in Figure \ref{fig:ablation}(a). A key observation is that, since the alignment strategy is weak, when done independently it does not lead to any performance gains. However, the proposed optimization provides significant improvement over even a joint optimization strategy.

\noindent\textbf{Convergence of $\mathrm{\Phi}$}:
Through Figure \ref{fig:ablation}(b) we report the training behavior of the alignment matrix $\mathrm{\Phi}$ returned by the auxiliary network $\mathcal{A}_{\Phi}$. 
While the red curve in Figure \ref{fig:ablation}(b) indicates the change in $\mathrm{\Phi}$ across iterations indexed by $t$ w.r.t the closed form solution $\mathrm{\Phi}^{init}$ obtained in \eqref{eq:globalsoln}, the blue curve represents the successive difference in $\mathrm{\Phi}$ across iterations. As expected, the estimate for $\mathrm{\Phi}$ changes non-trivially from $\mathrm{\Phi}^{init}$, eventually converging to a solution that leads to maximal classification performance. Note, in all our experiments, we find that the $\mathrm{\Phi}$ returned by the auxiliary network is always a well-conditioned, full rank matrix. 

% however as the algorithm reaches convergence w.r.t to target test accuracy, there is a reduction in rate of change as evidenced from the blue curve. 

\begin{figure}
    \centering
    \includegraphics[width= 0.9\columnwidth]{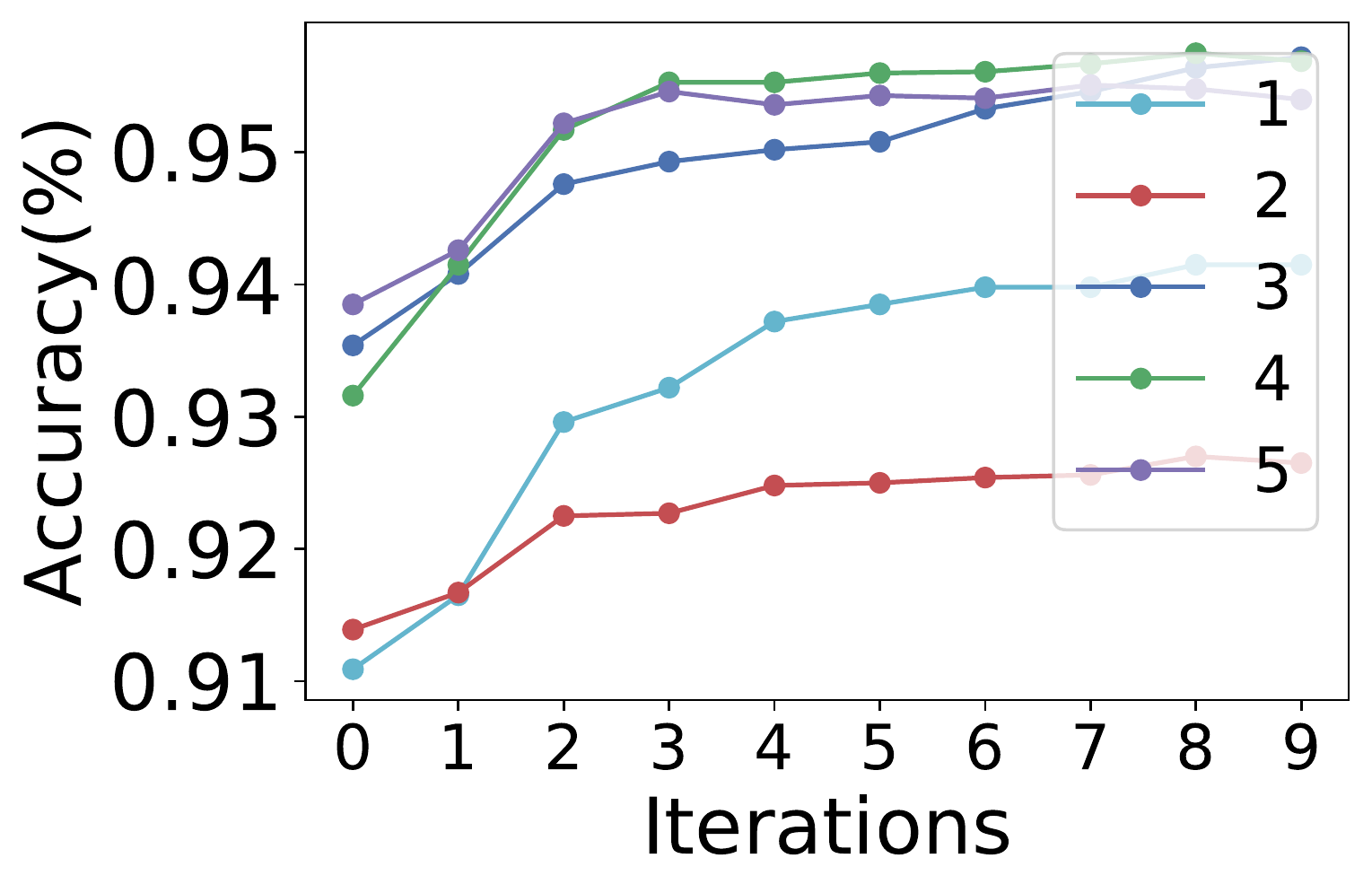}
    \caption{\textit{SVHN $\rightarrow$ MNIST DA task} - Effect of using an ensemble of target subspaces on the performance of SALT.}
    \label{fig:more_subspaces}
\end{figure}

\subsection{Digits classification}

\noindent\textbf{Datasets:} We consider three data sources for the digits classification task: USPS \cite{hull1994database}, MNIST \cite{lecun2010mnist}, and the Street View House Numbers (SVHN) \cite{netzer2011reading} datasets.
Each of these datasets have 10 categories (digits from 0-9). 
% The USPS dataset contains $7291$ training and $2007$ test grayscale images of handwritten images, each one of size $16 \times 16$ pixels.
% The MNIST dataset contains $60,000$ training and $10,000$ testing grayscale images of size $28 \times 28$. The SVHN dataset contains house numbers extracted from Google Street View images. This dataset contains $73,212$ training images, and $26,032$ testing images of size $32 \times 32 \times 3$. 
We perform the following three experiments in this task.
a) MNIST $\rightarrow$ USPS, b) USPS $\rightarrow$ MNIST, and c) SVHN $\rightarrow$ MNIST and report the accuracies on the standard target test sets.

% \begin{figure*}[h]
%     \centering
%     \subfloat[Ablation study]{{\includegraphics[width=0.45\linewidth]{figures/image_clef_ablation.pdf} }}%
%     \qquad
%   % \newline
%     \centering
%     \subfloat[Using multiple target subspaces]{{\includegraphics[width=0.45\linewidth]{figures/more_target_subspaces_SVHN_MNIST.pdf} }}%
%     % \includegraphics[width=0.95\linewidth]{figures/tsne.pdf} 
%     \caption{(a) Ablating different components in the proposed method against adaptation performance on the ImageCLEF dataset. See text in sec \ref{sec-imageclef} for notation.
%     (b) Effect of using multiple target subspaces in SALT on the SVHN-MNIST DA task.}%
%     \label{fig:ablation}%
%     %\vspace{-0.3in}
% \end{figure*}

\begin{table}[t]

    \label{tab:VISDA_Digits}
    \begin{center}
    \vspace{0.1in}

    %\setlength{\tabcolsep}{10pt} % Default value: 6pt
    
   % \resizebox{\columnwidth}{
    \subfloat[Digits datastets]{
    \renewcommand*{\arraystretch}{1.2}
    \begin{tabular}{cp{0.5in}p{0.4in}p{0.4in}}
    
    \hline
    Method& \small{MNIST$\rightarrow$ USPS} & \small{USPS$\rightarrow$ MNIST} & \small{SVHN$\rightarrow$ MNIST} \\
    \hline
    No Adaptation & 94.8 & 49 & 60.7 \\

    DeepCoRAL~\cite{deepcoral} & 89.3& 91.5& 59.6 \\
    MMD~\cite{MMD} & 88.5& 73.5& 64.8\\
    DANN~\cite{cite:JMLR16RevGrad} & 95.7& 90.0& 70.8\\
    ADDA~\cite{tzeng2017adversarial} & 92.4 & 93.8 & 76.0\\
    DeepJdot~\cite{bhushan2018deepjdot} & 95.6 & 96.0 & \textbf{96.7} \\
    CyCADA~\cite{CYCADA} & 95.6 & \mbox{\bf \em 96.5} & 90.9 \\
    UNIT~\cite{UNIT} & \mbox{\bf \em 95.9} &93.5& 90.5 \\
    GenToAdapt~\cite{GTA} & 95.3& 90.8 &92.4 \\
    \hline
    SALT & \textbf{96.2} & \textbf{97.4} & \mbox{\bf \em 95.6} \\
      \hline
  
    \end{tabular}%}
   \label{table:digits}
    }
    \newline
    \subfloat[VISDA-2017]{
    \renewcommand*{\arraystretch}{1.2}
    \label{table:visda}
     \begin{tabular}{cp{1.5cm}}
     \hline
        Method& Average Accuracy   \\\hline
        No Adaptation & 54.2\\
         JAN~\cite{JAN} & 61.6 \\
        CDAN~\cite{CDAN} & \mbox{\bf \em 70.2}\\
        \hline
       SALT &  \textbf{76.3}\\
        \hline
        \end{tabular}
          \label{table:visda}
    }
    \caption{\small{Performance of the proposed method on Digits and VISDA datasets. We highlight the best performing technique in {\bf bold}, and the second best in {\bf \em bold italic}.}}
    \label{table:digits}
    \end{center}
    \vspace{-0.2in}
    \end{table}

\noindent\textbf{Model:} The model used for all the $3$ tasks is based on the architecture from~\cite{bhushan2018deepjdot}, which is comprised of six $3 \times 3$ convolutional layers containing $\{32, 32, 64, 64, 128, 128\}$ filters with ReLU activations and two fully-connected layers of $128$ and $10$ (number of classes) hidden units. The Adam optimizer ($lr = 2e^{-4}$) was used to update the  model using a mini-batch size of $512$ for the two domains. We compare our results with a number of state-of-the-art domain adaptation methods and the results are shown in Table \ref{table:digits}. SALT achieves the highest accuracy averaged across all three digits datasets, beating state-of-the-art in two out of three cases, and second highest marginally below DeepJDOT \cite{bhushan2018deepjdot}, in the case of SVHN $\rightarrow$ MNIST. 

With one of the tasks in this dataset, we also study the effect of using multiple subspaces on the classification performance. As discussed earlier, allowing multiple target subspaces increases the complexity of the auxiliary task. As showed in Figure \ref{fig:more_subspaces}, with the SVHN $\rightarrow$ MNIST DA task, using 3 or more subspaces leads to significant performance gains. However, we found that increasing it further did not lead to additional improvements.

%\footnote{\url{http://ai.bu.edu/visda-2017/}}
%\vspace{-0.1in}
\subsection{VisDA-2017}
\noindent\textbf{Dataset}:
VisDA-2017 
is a difficult simulation-to-realworld dataset with two highly distinct domains: 
\textit{Synthetic}, renderings of 3D models from different angles and with different lightning conditions; \textit{Real} which are natural images. This dataset contains over 280K images across 12 classes.

\noindent\textbf{Model}:
Given the dataset complexity, we choose the pretrained ResNet-152~\cite{ResNet} as our feature extractor and as in previous case, we fine tune it to obtain the $2048$-dimensional features and the subspace dimension is $800$. 
The classifier and subspace alignment networks are trained with the same hyper-parameters as in Section \ref{sec-imageclef}.  From Table \ref{table:visda}, it can be seen that our model comprehensively outperforms the results reported so far in the literature, by six percentage points.

%Since our model is based on \cite{bhushan2018deepjdot}  and it re-implements the following methods with respect to the model used, 
% We report the numbers of \cite{deepcoral,MMD,cite:JMLR16RevGrad,tzeng2017adversarial} directly from \cite{bhushan2018deepjdot}. For other methods, we report numbers directly from the respective papers.

% Even though we do not use complex adversarial based methods, we achieve accuracies comparable or higher than the state-of-the art methods. 

\begin{figure*}[t]
    \centering
    \subfloat[No adaptation]{{\includegraphics[trim={5cm 5cm 5cm 5cm},clip,width=0.35\linewidth]{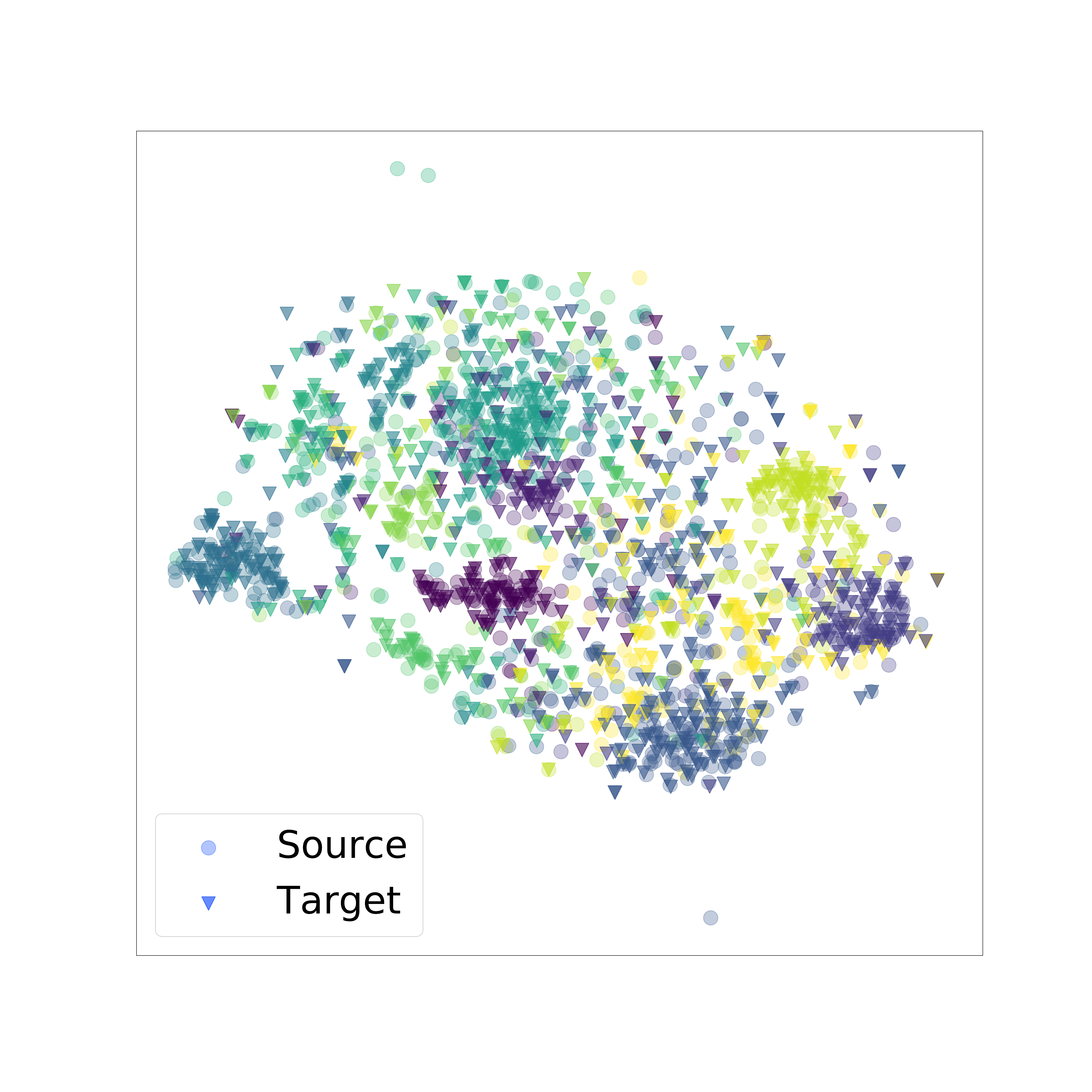} }}%
    \qquad
    \subfloat[After adaptation]{{\includegraphics[trim={5cm 5cm 5cm 5cm},clip,width=0.35\linewidth]{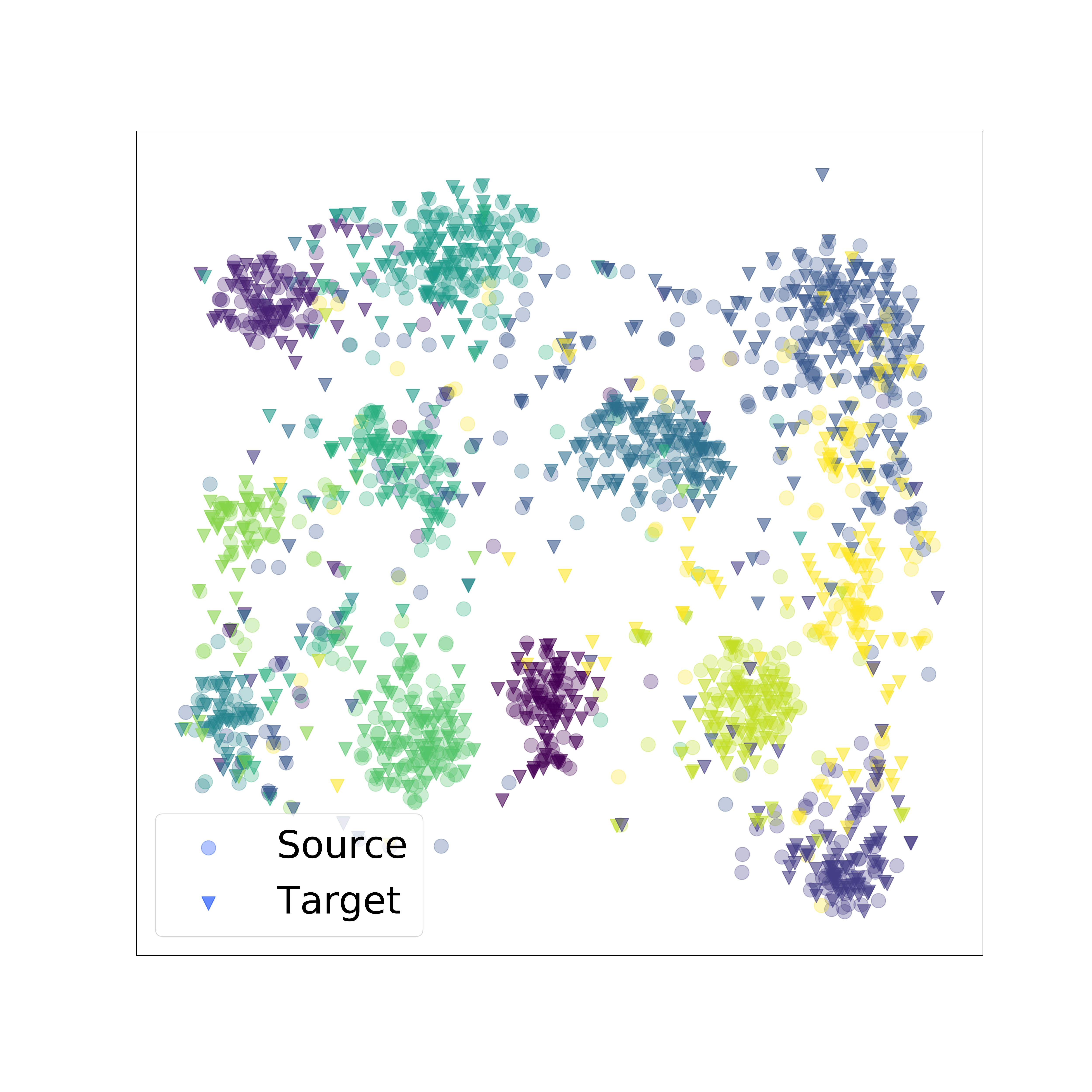} }}%
    \caption{\textit{VisDA-2017} - Visualizing the adaptation across source and target domains using t-SNE \cite{maaten2008visualizing}. We observe improved alignment between the class boundaries of the source and target domains.}%
    \label{fig:tsne}%
\end{figure*}

\subsection{Office-Home}

\noindent\textbf{Datasets}: This challenging dataset~\cite{cite:CVPR17DHN} contains 15,500 images in 65 classes from office and home settings, forming $4$ extremely dissimilar domains: Artistic images (\textbf{Ar}), Clip Art (\textbf{Cl}), Product images (\textbf{Pr}), and Real-World images (\textbf{Rw}).

\noindent\textbf{Model}: Similar to Section \ref{sec-imageclef}, we fine tune a pre-trained ResNet-50 and obtain the $2048$-dimensional features, and build subspaces of $800$ dimensions. The classifier and auxiliary networks are trained with the same hyper-parameters as earlier. 
Comparisons to the state-of-the-art methods can be found in Table \ref{table:officehome}. We observe that while SALT consistently outperforms baseline methods including the recent DeepJdot~\cite{bhushan2018deepjdot}, with comparable performance to the highest reported -- CDAN \cite{CDAN} in terms of the average accuracy across all pairs of DA tasks (see Supplementary for detailed results).

%\begin{figure}[h]
%  \centering
%  \includegraphics[width=0.8\textwidth]{Dataset.pdf}
%  \caption{Example images of the Office-Home dataset (left) and VisDA-2017 dataset (right).}
%  \label{fig:data}
%\end{figure}

\begin{table}[h]

    \begin{center}
    %\setlength{\tabcolsep}{10pt} % Default value: 6pt
   % \renewcommand{\arraystretch}{1.3}
    %\resizebox{\columnwidth}{!}{%
  \begin{tabular}{cp{1.5cm}}
  \hline
Method &  Average Accuracy \\  \hline
No Adaptation & 59.7 \\  
 DeepJdot~\cite{bhushan2018deepjdot} &  50.6 \\
DAN~\cite{DAN}  & 56.3\\
DANN~\cite{cite:JMLR16RevGrad} & 57.6\\
JAN~\cite{JAN}  &58.3\\
CDAN~\cite{CDAN} &  \textbf{65.8} \\   
\hline
SALT  &\mbox{\bf \em 65.1} \\ \hline
\end{tabular}
%}
\caption{\small{Classification accuracy on Office-Home dataset. Best performance is shown in {\bf bold}, and the second best in {\bf \em bold italic}.}}
\label{table:officehome}
\end{center}
\vspace{-20pt}
\end{table}

 \subsection{SALT relaxes data requirements}
\begin{figure}[!htb]
    \centering
    \includegraphics[width = 0.9\columnwidth]{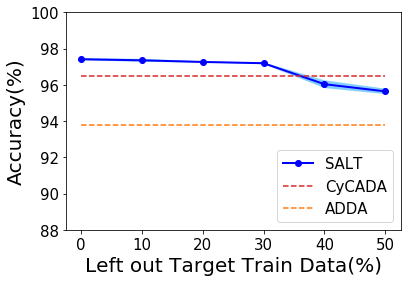}
    \caption{Performance on the target test-set with varying sizes of target training data on the USPS to MNIST adaptation task. The baseline results with ADDA and CyCADA were obtained using the entire target training data.}
    \label{fig:reduced_data}
    \vspace{-15pt}
\end{figure}

 Owing to its design simplicity, we surmise that \salt admits improved data efficiency. 
 To test this hypothesis, we evaluate \salt under the scenario where the amount of unlabeled target data is limited. While we perform SALT with varying target data sizes, we report accuracies on the full target test set. In particular, we consider the problem of adapting USPS $\rightarrow$ MNIST and the results from $3$ random trials are shown in Figure \ref{fig:reduced_data}. It can be seen that even with $30\%$ lesser training data in the target domain, SALT still outperforms state-of-the-art baselines that have access to the entire data. Further, the drop in performance even when operating at only 50\% of data is very low ($\approx$ 2\% points), thus evidencing that a simpler alignment strategy can reduce the data requirements while not compromising the performance.

 \section{Conclusions}
In this work, we present a principled and effective approach to tackle the problem of unsupervised domain adaptation, in the context of visual recognition. The proposed method -- \salt -- poses alignment as an auxiliary task to the primary task of maximizing performance on both source and target datasets. \salt proposes to solve domain alignment by utilizing gradients from the primary task. It utilizes an alternating optimization strategy, without refining the feature extractor, thus providing a venue for systematic control of domain alignment under the objective of generalizing to the target set.  
Through an extensive quantitative and qualitative evaluation, it is shown that \salt achieves comparable or sometimes higher performance than the state-of-the-art on multiple benchmarks. Future work includes extending the SALT methodology to newer tasks such as as semantic segmentation, open-set classification~\cite{Saito_2018_ECCV}, and image-to-image translation.

%\salt is generic, and can be used in conjunction with any other feature extractor. 

\label{Conclusion}
 
{\small
\bibliographystyle{ieee_fullname}
\bibliography{references}

\begin{thebibliography}{10}\itemsep=-1pt

\bibitem{Ben-David:2010:TLD:1745449.1745461}
Shai Ben-David, John Blitzer, Koby Crammer, Alex Kulesza, Fernando Pereira, and
  Jennifer~Wortman Vaughan.
\newblock A theory of learning from different domains.
\newblock {\em Mach. Learn.}, 79(1-2):151--175, May 2010.

\bibitem{bhushan2018deepjdot}
Bharath Bhushan~Damodaran, Benjamin Kellenberger, R{\'e}mi Flamary, Devis Tuia,
  and Nicolas Courty.
\newblock Deepjdot: Deep joint distribution optimal transport for unsupervised
  domain adaptation.
\newblock In {\em Proceedings of the European Conference on Computer Vision
  (ECCV)}, pages 447--463, 2018.

\bibitem{bousmalis2017unsupervised}
Konstantinos Bousmalis, Nathan Silberman, David Dohan, Dumitru Erhan, and Dilip
  Krishnan.
\newblock Unsupervised pixel-level domain adaptation with generative
  adversarial networks.
\newblock In {\em Proceedings of the IEEE conference on computer vision and
  pattern recognition}, pages 3722--3731, 2017.

\bibitem{Courty2017OptimalTF}
Nicolas Courty, R{\'e}mi Flamary, Devis Tuia, and Alain Rakotomamonjy.
\newblock Optimal transport for domain adaptation.
\newblock {\em IEEE Transactions on Pattern Analysis and Machine Intelligence},
  39:1853--1865, 2017.

\bibitem{Subspace_alignment}
Basura Fernando, Amaury Habrard, Marc Sebban, and Tinne Tuytelaars.
\newblock Unsupervised visual domain adaptation using subspace alignment.
\newblock In {\em {IEEE} International Conference on Computer Vision, {ICCV}
  2013, Sydney, Australia, December 1-8, 2013\/} \cite{Subspace_alignment},
  pages 2960--2967.

\bibitem{Fernando2014SubspaceAF}
Basura Fernando, Amaury Habrard, Marc Sebban, and Tinne Tuytelaars.
\newblock Subspace alignment for domain adaptation.
\newblock {\em CoRR}, abs/1409.5241, 2014.

\bibitem{maml}
Chelsea Finn, Pieter Abbeel, and Sergey Levine.
\newblock Model-agnostic meta-learning for fast adaptation of deep networks.
\newblock In {\em Proceedings of the 34th International Conference on Machine
  Learning-Volume 70}, pages 1126--1135. JMLR. org, 2017.

\bibitem{FrenchMF18}
Geoffrey French, Michal Mackiewicz, and Mark~H. Fisher.
\newblock Self-ensembling for visual domain adaptation.
\newblock In {\em The 6th International Conference on Learning Representations
  {(ICLR)}}, 2018.

\bibitem{cite:ICML15RevGrad}
Yaroslav Ganin and Victor Lempitsky.
\newblock Unsupervised domain adaptation by backpropagation.
\newblock {\em arXiv preprint arXiv:1409.7495}, 2014.

\bibitem{cite:JMLR16RevGrad}
Yaroslav Ganin, Evgeniya Ustinova, Hana Ajakan, Pascal Germain, Hugo
  Larochelle, Fran{\c{c}}ois Laviolette, Mario Marchand, and Victor Lempitsky.
\newblock Domain-adversarial training of neural networks.
\newblock {\em The Journal of Machine Learning Research}, 17(1):2096--2030,
  2016.

\bibitem{Gong2012GeodesicFK}
Boqing Gong, Yuan Shi, Fei Sha, and Kristen Grauman.
\newblock Geodesic flow kernel for unsupervised domain adaptation.
\newblock {\em 2012 IEEE Conference on Computer Vision and Pattern
  Recognition}, pages 2066--2073, 2012.

\bibitem{cite:NIPS14GAN}
Ian Goodfellow, Jean Pouget-Abadie, Mehdi Mirza, Bing Xu, David Warde-Farley,
  Sherjil Ozair, Aaron Courville, and Yoshua Bengio.
\newblock Generative adversarial nets.
\newblock pages 2672--2680, 2014.

\bibitem{Gopalan2011DomainAF}
Raghuraman Gopalan, Ruonan Li, and Rama Chellappa.
\newblock Domain adaptation for object recognition: An unsupervised approach.
\newblock {\em 2011 International Conference on Computer Vision}, pages
  999--1006, 2011.

\bibitem{ResNet}
Kaiming He, Xiangyu Zhang, Shaoqing Ren, and Jian Sun.
\newblock Deep residual learning for image recognition.
\newblock pages 770--778, 2016.

\bibitem{judyiclr13}
Judy Hoffman, Erik Rodner, Jeff Donahue, Kate Saenko, and Trevor Darrell.
\newblock Efficient learning of domain-invariant image representations.
\newblock In {\em 1st International Conference on Learning Representations,
  {ICLR} 2013, Scottsdale, Arizona, USA, May 2-4, 2013, Conference Track
  Proceedings}, 2013.

\bibitem{CYCADA}
Judy Hoffman, Eric Tzeng, Taesung Park, Jun{-}Yan Zhu, Phillip Isola, Kate
  Saenko, Alexei~A. Efros, and Trevor Darrell.
\newblock Cycada: Cycle-consistent adversarial domain adaptation.
\newblock In {\em Proceedings of the 35th International Conference on Machine
  Learning, {ICML} 2018, Stockholmsm{\"{a}}ssan, Stockholm, Sweden, July 10-15,
  2018\/} \cite{CYCADA}, pages 1994--2003.

\bibitem{hull1994database}
Jonathan~J. Hull.
\newblock A database for handwritten text recognition research.
\newblock {\em IEEE Transactions on pattern analysis and machine intelligence},
  16(5):550--554, 1994.

\bibitem{cite:CVPR17pix2pix}
Phillip Isola, Jun-Yan Zhu, Tinghui Zhou, and Alexei~A Efros.
\newblock Image-to-image translation with conditional adversarial networks.
\newblock pages 1125--1134, 2017.

\bibitem{JaderbergMCSLSK17}
Max Jaderberg, Volodymyr Mnih, Wojciech~Marian Czarnecki, Tom Schaul, Joel~Z.
  Leibo, David Silver, and Koray Kavukcuoglu.
\newblock Reinforcement learning with unsupervised auxiliary tasks.
\newblock In {\em 5th International Conference on Learning Representations,
  {ICLR} 2017, Toulon, France, April 24-26, 2017, Conference Track
  Proceedings}, 2017.

\bibitem{ADAM}
Diederik~P Kingma and Jimmy Ba.
\newblock Adam: A method for stochastic optimization.
\newblock {\em arXiv preprint arXiv:1412.6980}, 2014.

\bibitem{koch2015siamese}
Gregory Koch, Richard Zemel, and Ruslan Salakhutdinov.
\newblock Siamese neural networks for one-shot image recognition.
\newblock In {\em ICML deep learning workshop}, volume~2, 2015.

\bibitem{lecun2010mnist}
Yann LeCun, Corinna Cortes, and CJ Burges.
\newblock Mnist handwritten digit database.
\newblock {\em AT\&T Labs [Online]. Available: http://yann. lecun.
  com/exdb/mnist}, 2:18, 2010.

\bibitem{Liebel}
Lukas Liebel and Marco K{\"{o}}rner.
\newblock Auxiliary tasks in multi-task learning.
\newblock {\em CoRR}, abs/1805.06334, 2018.

\bibitem{UNIT}
Ming-Yu Liu, Thomas Breuel, and Jan Kautz.
\newblock Unsupervised image-to-image translation networks.
\newblock In I. Guyon, U.~V. Luxburg, S. Bengio, H. Wallach, R. Fergus, S.
  Vishwanathan, and R. Garnett, editors, {\em Advances in Neural Information
  Processing Systems}, pages 700--708. Curran Associates, Inc., 2017.

\bibitem{liu2016coupled}
Ming-Yu Liu and Oncel Tuzel.
\newblock Coupled generative adversarial networks.
\newblock In {\em Advances in neural information processing systems}, pages
  469--477, 2016.

\bibitem{metaauxillary}
Shikun Liu, Andrew~J Davison, and Edward Johns.
\newblock Self-supervised generalisation with meta auxiliary learning.
\newblock {\em arXiv preprint arXiv:1901.08933}, 2019.

\bibitem{DAN}
Mingsheng Long, Yue Cao, Jianmin Wang, and Michael~I Jordan.
\newblock Learning transferable features with deep adaptation networks.
\newblock {\em arXiv preprint arXiv:1502.02791}, 2015.

\bibitem{MMD}
Mingsheng Long, Yue Cao, Jianmin Wang, and Michael~I. Jordan.
\newblock Learning transferable features with deep adaptation networks.
\newblock In {\em Proceedings of the 32Nd International Conference on
  International Conference on Machine Learning - Volume 37}, ICML'15, pages
  97--105. JMLR.org, 2015.

\bibitem{CDAN}
Mingsheng Long, Zhangjie Cao, Jianmin Wang, and Michael~I. Jordan.
\newblock Conditional adversarial domain adaptation.
\newblock In {\em Advances in Neural Information Processing Systems
  {(NeurIPS)}}, pages 1647--1657, 2018.

\bibitem{JAN}
Mingsheng Long, Han Zhu, Jianmin Wang, and Michael~I Jordan.
\newblock Deep transfer learning with joint adaptation networks.
\newblock In {\em Proceedings of the 34th International Conference on Machine
  Learning-Volume 70}, pages 2208--2217. JMLR. org, 2017.

\bibitem{maaten2008visualizing}
Laurens van~der Maaten and Geoffrey Hinton.
\newblock Visualizing data using t-sne.
\newblock {\em Journal of machine learning research}, 9(Nov):2579--2605, 2008.

\bibitem{MunkhdalaiY17}
Tsendsuren Munkhdalai and Hong Yu.
\newblock Meta networks.
\newblock In {\em Proceedings of the 34th International Conference on Machine
  Learning, {ICML} 2017, Sydney, NSW, Australia, 6-11 August 2017}, pages
  2554--2563, 2017.

\bibitem{netzer2011reading}
Yuval Netzer, Tao Wang, Adam Coates, Alessandro Bissacco, Bo Wu, and Andrew~Y
  Ng.
\newblock Reading digits in natural images with unsupervised feature learning.
\newblock 2011.

\bibitem{pan2010domain}
Sinno~Jialin Pan, Ivor~W Tsang, James~T Kwok, and Qiang Yang.
\newblock Domain adaptation via transfer component analysis.
\newblock {\em IEEE Transactions on Neural Networks}, 22(2):199--210, 2010.

\bibitem{paszke2017automatic}
Adam Paszke, Sam Gross, Soumith Chintala, Gregory Chanan, Edward Yang, Zachary
  DeVito, Zeming Lin, Alban Desmaison, Luca Antiga, and Adam Lerer.
\newblock Automatic differentiation in pytorch.
\newblock In {\em NIPS-W}, 2017.

\bibitem{RaviL17}
Sachin Ravi and Hugo Larochelle.
\newblock Optimization as a model for few-shot learning.
\newblock In {\em 5th International Conference on Learning Representations,
  {ICLR} 2017, Toulon, France, April 24-26, 2017, Conference Track
  Proceedings}, 2017.

\bibitem{russakovsky2015imagenet}
Olga Russakovsky, Jia Deng, Hao Su, Jonathan Krause, Sanjeev Satheesh, Sean Ma,
  Zhiheng Huang, Andrej Karpathy, Aditya Khosla, Michael Bernstein, et~al.
\newblock Imagenet large scale visual recognition challenge.
\newblock {\em International journal of computer vision}, 115(3):211--252,
  2015.

\bibitem{Saenko2010AdaptingVC}
Kate Saenko, Brian Kulis, Mario Fritz, and Trevor Darrell.
\newblock Adapting visual category models to new domains.
\newblock In {\em ECCV}, 2010.

\bibitem{Saito_2018_ECCV}
Kuniaki Saito, Shohei Yamamoto, Yoshitaka Ushiku, and Tatsuya Harada.
\newblock Open set domain adaptation by backpropagation.
\newblock In {\em The European Conference on Computer Vision (ECCV)}, September
  2018.

\bibitem{GTA}
Swami Sankaranarayanan, Yogesh Balaji, Carlos~D Castillo, and Rama Chellappa.
\newblock Generate to adapt: Aligning domains using generative adversarial
  networks.
\newblock In {\em Proceedings of the IEEE Conference on Computer Vision and
  Pattern Recognition}, pages 8503--8512, 2018.

\bibitem{Santoro:2016:MMN:3045390.3045585}
Adam Santoro, Sergey Bartunov, Matthew Botvinick, Daan Wierstra, and Timothy
  Lillicrap.
\newblock Meta-learning with memory-augmented neural networks.
\newblock In {\em Proceedings of the 33rd International Conference on
  International Conference on Machine Learning - Volume 48}, ICML'16, pages
  1842--1850. JMLR.org, 2016.

\bibitem{ShrivastavaWACV2014}
Ashish Shrivastava, Sumit Shekhar, and Vishal~M Patel.
\newblock Unsupervised domain adaptation using parallel transport on grassmann
  manifold.
\newblock In {\em Applications of Computer Vision (WACV), 2014 IEEE Winter
  Conference on}, pages 277--284. IEEE, 2014.

\bibitem{shu2018a}
Rui Shu, Hung Bui, Hirokazu Narui, and Stefano Ermon.
\newblock A {DIRT}-t approach to unsupervised domain adaptation.
\newblock In {\em International Conference on Learning Representations}, 2018.

\bibitem{sun2017correlation}
Baochen Sun, Jiashi Feng, and Kate Saenko.
\newblock Correlation alignment for unsupervised domain adaptation.
\newblock In {\em Domain Adaptation in Computer Vision Applications}, pages
  153--171. Springer, 2017.

\bibitem{sun2015subspace}
Baochen Sun and Kate Saenko.
\newblock Subspace distribution alignment for unsupervised domain adaptation.
\newblock In {\em BMVC}, pages 24--1, 2015.

\bibitem{deepcoral}
Baochen Sun and Kate Saenko.
\newblock Deep coral: Correlation alignment for deep domain adaptation.
\newblock In {\em European Conference on Computer Vision}, pages 443--450.
  Springer, 2016.

\bibitem{thopalli2019multiple}
Kowshik Thopalli, Rushil Anirudh, Jayaraman~J Thiagarajan, and Pavan Turaga.
\newblock Multiple subspace alignment improves domain adaptation.
\newblock In {\em ICASSP 2019-2019 IEEE International Conference on Acoustics,
  Speech and Signal Processing (ICASSP)}, pages 3552--3556. IEEE, 2019.

\bibitem{ToshniwalTLL17}
Shubham Toshniwal, Hao Tang, Liang Lu, and Karen Livescu.
\newblock Multitask learning with low-level auxiliary tasks for encoder-decoder
  based speech recognition.
\newblock In {\em Interspeech 2017, 18th Annual Conference of the International
  Speech Communication Association, Stockholm, Sweden, August 20-24, 2017},
  pages 3532--3536, 2017.

\bibitem{tzeng2017adversarial}
Eric Tzeng, Judy Hoffman, Kate Saenko, and Trevor Darrell.
\newblock Adversarial discriminative domain adaptation.
\newblock In {\em Computer Vision and Pattern Recognition (CVPR)}, volume~1,
  page~4, 2017.

\bibitem{cite:CVPR17DHN}
Hemanth Venkateswara, Jose Eusebio, Shayok Chakraborty, and Sethuraman
  Panchanathan.
\newblock Deep hashing network for unsupervised domain adaptation.
\newblock pages 5018--5027, 2017.

\bibitem{vinyals2016matching}
Oriol Vinyals, Charles Blundell, Timothy Lillicrap, Daan Wierstra, et~al.
\newblock Matching networks for one shot learning.
\newblock In {\em Advances in neural information processing systems}, pages
  3630--3638, 2016.

\end{thebibliography}
}

\end{document}

%%%%%%%%% TITLE
\title{Supplement to ``\salt: Subspace Alignment as an Auxiliary Learning Task for Domain Adaptation''}

\maketitle
\appendix
\section{Results on Office-Home Dataset }
While we reported aggregate statistics in Table 3 of the main paper for the Office-Home Dataset~\cite{cite:CVPR17DHN}, here, in Table \ref{table:officehome_sup} we report detailed performance across all pairs of DA tasks for this dataset.
From Table \ref{table:officehome_sup} we observe that while SALT consistently outperforms baseline methods including the recent DeepJdot~\cite{bhushan2018deepjdot}, it also achieves comparable performance to the highest reported -- CDAN \cite{CDAN} in all DA tasks. 
\begin{table*}[bp]

    \begin{center}

    %\setlength{\tabcolsep}{10pt} % Default value: 6pt
  % \renewcommand{\arraystretch}{1.3}
    %\resizebox{\columnwidth}{!}{%

  \begin{tabular}{cp{0.2in}p{0.2in}p{0.2in}p{0.2in}p{0.2in}p{0.2in}p{0.2in}p{0.2in}p{0.2in}p{0.2in}p{0.2in}p{0.2in}p{0.2in}}
  \hline
Method & Ar $\rightarrow$ Cl & Ar $\rightarrow$ Pr & Ar $\rightarrow$ RW & Cl $\rightarrow$ Ar & Cl $\rightarrow$ Pr & Cl $\rightarrow$ Rw & Pr $\rightarrow$ Ar & Pr $\rightarrow$ Cl & Pr $\rightarrow$ Rw & Rw $\rightarrow$ Ar & Rw $\rightarrow$ Cl & Rw $\rightarrow$ Pr & Avg \\  \hline
No Adaptation & 44.6 & 62.7 & 72.0 & 52.1 & 62.7 & 65.1 & 52.9 & 43.0 & 73.9 & 63.7 & 45.8 & 77.3 & 59.7 \\  
 DeepJdot~\cite{bhushan2018deepjdot} & 39.7 & 50.4 & 62.5 & 39.5 & 54.4 & 53.2 & 36.7 & 39.2 & 63.5 & 52.3 & 45.4 & 70.5 & 50.6 \\
DAN~\cite{DAN}  & 43.6 & 57.0  & 67.9 & 45.8 & 56.5 & 60.4 & 44.0 & 43.6 & 67.7 & 63.1 & 51.5 & 74.3 & 56.3\\
DANN~\cite{cite:JMLR16RevGrad} & 45.6 & 59.3  &70.1 &47.0 & 58.5  &60.9 & 46.1& 43.7& 68.5& 63.2& 51.8& 76.8& 57.6\\
JAN~\cite{JAN}  &45.9 &61.2 &68.9 &50.4 &59.7 &61.0 &45.8 &43.4 &70.3 &63.9 &52.4 &76.8 &58.3\\
CDAN~\cite{CDAN} & \textbf{50.7} & \textbf{70.6} & \textbf{76} & \mbox{\bf \em 57.6} & \textbf{70} & \mbox{\bf \em 70} & \mbox{\bf \em 57.4} & \textbf{50.9} & \textbf{77.3} & \textbf{70.9} & \textbf{56.7} & \textbf{81.6} & \textbf{65.8} \\   
\hline
SALT  &\mbox{\bf \em  49.9} & \mbox{\bf \em 68.6} & \mbox{\bf \em 74.68} & \textbf{59.9} & \mbox{\bf \em 68.92} & \textbf{71.82} & \textbf{58.12} & \mbox{\bf \em 49.4} & \textbf{77.3} & \mbox{\bf \em 68.7} & \mbox{\bf \em 54.82} & \mbox{\bf \em 78.92} & \mbox{\bf \em 65.1} \\ \hline
\end{tabular}
%}
\caption{Classification accuracy on Office-Home dataset. Best performance is shown in {\bf bold}, and the second best in {\bf \em bold italic}.}
\label{table:officehome_sup}
\end{center}

\end{table*}

{\small
\bibliographystyle{ieee_fullname}
\bibliography{references}
}